\documentclass{article}
\usepackage[final,sglblindworkshop]{neurips_2026}
\usepackage[utf8]{inputenc} % allow utf-8 input
\usepackage[T1]{fontenc}    % use 8-bit T1 fonts
\usepackage{hyperref}       % hyperlinks
\usepackage{url}            % simple URL typesetting
\usepackage{booktabs}       % professional-quality tables
\usepackage{amsfonts}       % blackboard math symbols
\usepackage{nicefrac}       % compact symbols for 1/2, etc.
\usepackage{microtype}      % microtypography
\usepackage{xcolor}         % colors
\usepackage{graphicx}

\renewcommand{\acksection}{\section*{Acknowledgments}}

\workshoptitle{Interpretability for Discovery} 

\title{Mechanistic Interpretability Needs a Logic of Inference}

\author{
  \textbf{Zhimin Hu}$^{1}$\thanks{Correspondence to: Zhimin Hu $<$\texttt{zhu41@gatech.edu}$>$} \quad \textbf{Chen Jiang}$^{2}$ \quad \textbf{Hanbo Xie}$^{1}$ \quad \textbf{N.Apurva Ratan Murty}$^{1}$ \quad \textbf{Sashank Varma}$^{1}$ \\
  \vspace{1ex} \\
  $^{1}$ Georgia Tech \\
  $^{2}$ McGill University\\
}

\begin{document}

\maketitle

\begin{abstract}
Neural networks are increasingly studied not merely as predictive systems, but as complex systems whose internal computations are themselves objects of scientific inquiry. Recent progress in mechanistic interpretability (MI) has made it possible to identify neurons, features, circuits, population structures, and dynamical patterns associated with model behavior. As these tools improve, attention is shifting from what can be observed inside a model to how those observations support mechanistic explanations. We examine the parallel efforts of neuroscience and MI to address this question and identify recurring inferential bottlenecks when inferring mechanisms from complex neural systems. We propose five principles and organize them into a recursive logic for a more rigorous procedure of mechanistic inference. These principles provide a framework that supports scientific discovery in two senses: it helps turn local interpretability findings into cumulative knowledge of mechanisms within models, and it makes those mechanisms sufficiently precise to generate testable hypotheses about domains beyond the models. Evidence beyond MI is ultimately required to establish what model mechanisms reveal about the world.
\end{abstract}

\section{Introduction}

Mechanistic interpretability (MI) research for artificial neural networks (ANNs) has generated remarkable advances in new methodologies over the past decades. These techniques are increasingly used to study internal representations and computations. However, the success of these observational tools has made the gap between seeing a feature and understanding the computation larger and more subtle. In earlier studies, researchers were tempted to conclude from probing that information is recoverable from a representation without showing that the model actually uses that information in its computations. Likewise, interventions have been used to establish that a component is causally relevant without identifying the computational role that component plays \citep{sharkey2025open}. Recent works have attempted to address these concerns through high-level conceptual frameworks \citep{vilas2024position} or localized practices \citep{geiger2025causal,joshi2026position}, but a general logic for moving from interpretability results to formal mechanistic claims remains missing.

We address this gap by examining the parallel efforts of neuroscientists to understand the brain and MI researchers to understand ANNs. Neuroscience has repeatedly confronted the same problem about what can be inferred from different kinds of observations about a neural system. Over decades, broad questions about neural mechanisms were factorized into more precise questions that separate what a component represents from how information is organized across populations, transformed over stages, picked up by downstream neurons, and preserved across different implementations\citep{pouget2000information, quian2009extracting,kriegeskorte2019interpreting, diedrichsen2017representational}. This thinking revealed a fundamental inferential bottleneck of connecting one class of evidence to the next when dissecting complex neural systems. We coalesce these insights into five principles for `mechanistic' inference. They concern identifying the appropriate explanatory object (P1), determining whether represented information is used in computation (P2), broadening from causal relevance to causal role (P3), moving from component role to organization (P4), and testing the invariance among levels of analysis and opening up new mechanistic questions (P5). Measurement conditions each of these inferences (P0).

Our principles support MI as a science of discovery in two related senses. First, when neural networks are the object of study, the principles provide a recursive logic for turning local interpretability findings into more constrained and cumulative accounts of model mechanisms. Second, models trained on scientific data can be analyzed mechanistically to produce knowledge beyond themselves. Our framework explains how MI can identify precise candidate mechanisms, which can serve as sources of precise hypotheses and discriminating predictions about that domain.

\begin{figure*}[t]
\centering
\includegraphics[width=\textwidth]{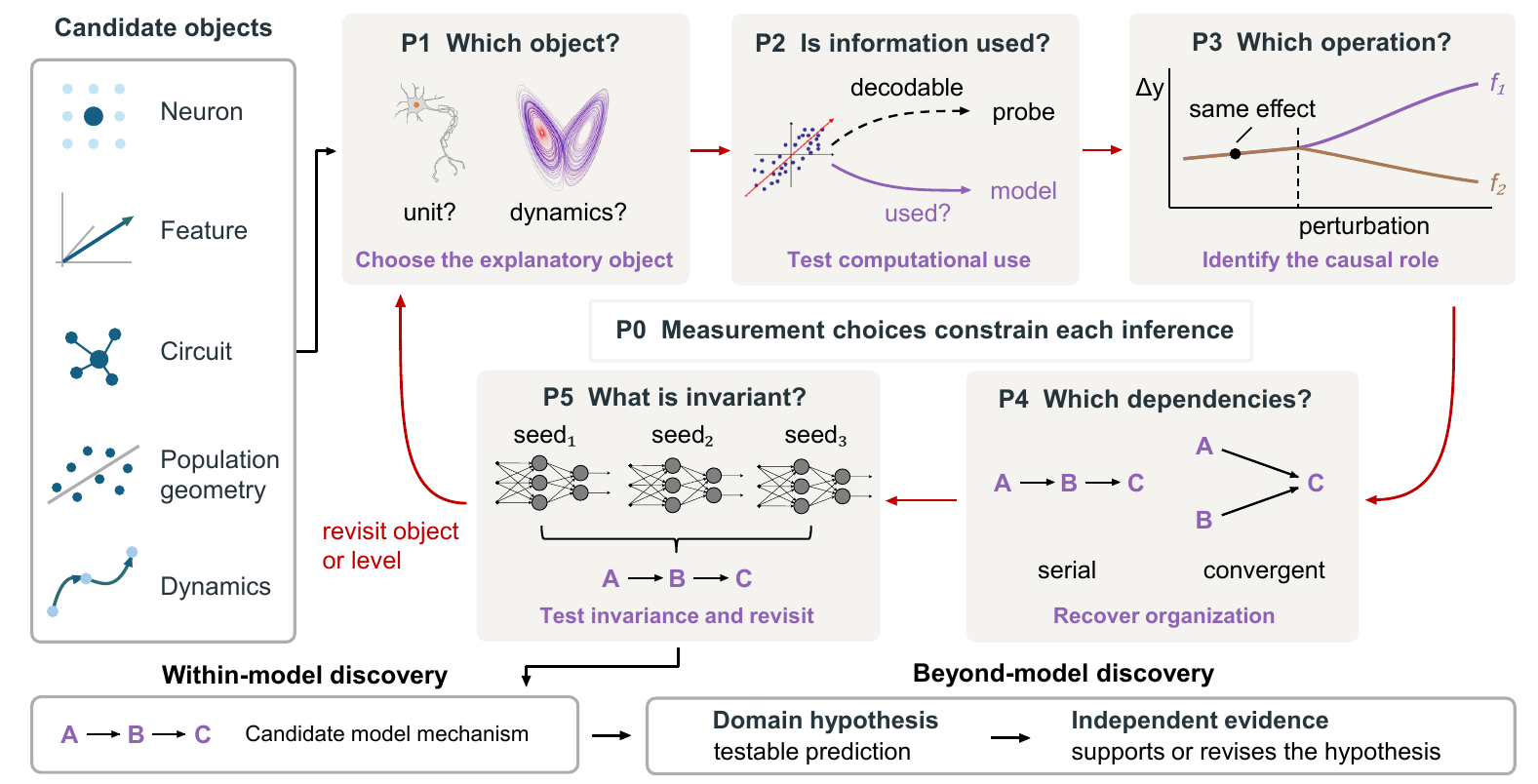}
\caption{\textbf{A recursive logic for mechanistic inference.}
For a given neural system with behavior of interest, it can be investigated by different candidate objects of study: neurons, features, circuits, population geometry, and dynamics. We propose five principles to organize mechanistic research as a recursive process in which evidence informs further questions and revisions to earlier claims. An individual study may address a subset of these questions and revisit them as needed. Red arrows connect these five inferential questions: which object explains the phenomenon (P1), whether represented information is used in computation (P2), what computational role it plays (P3), how roles are organized (P4), and what remains invariant across conditions or implementations (P5; illustrated here across random seeds). Measurement choices constrain every inference (P0). A, B, and C denote computational roles. This process supports the discovery of candidate model mechanisms (within-model discovery), which can generate domain hypotheses and establish their plausibility beyond the model requires independent evidence (beyond-model discovery).}
\label{fig:main}
\end{figure*}

\section{A Brief Tour of Mechanistic Interpretability}
\label{section2}

MI refers to a set of techniques for succinctly explaining the behavior of neural networks by reference to their internal representations and computations. These objects of explanation are the `mechanisms'. MI approaches differ in what they treat as the object of explanation.

Early work often focused on individual neurons or units. For example, \citet{simonyan2013deep, olah2017feature, olah2018the} identified units whose activations aligned with human-interpretable visual concepts. Such findings are straightforward; however, neural representations may not always align with individual neurons \citep{rigotti2013importance}. The possibility of distributed representations thus motivated the search for explanatory objects beyond individual neurons \citep{morcos2018importance}.

A newer approach is to recover alternative feature bases. Sparse autoencoders (SAEs) learn sparse decompositions of model activations and can identify features that are more interpretable than individual neurons \citep{huben2024sae}. Related approaches identify feature directions that can be intervened to steer model behaviors \citep{li2023inferencetime}. However, such features do not constitute a unique decomposition of model activity \citep{leask2025sparse}; as mechanisms, they are underdetermined.

Circuit-level approaches instead ask how multiple components interact to implement a computation. For example, \citet{wang2023interpretability} identified a circuit of interacting attention heads underlying identification of the indirect objects of clauses. More recent work builds circuits over learned sparse features rather than raw model components, illustrating the insights offered by combining feature- and circuit-level descriptions \citep{marks2025sparse}.

One level up in complexity are population-level analyses, which characterize the geometry and low-dimensional structure of activity across many units \citep{chung2021geo,hu2026language,feng2024how}. For recurrent or temporally extended computation, the relevant mechanism may instead be described as trajectories through a dynamical state space \citep{sussillo2013dynamics}.

These approaches are not mutually exclusive, but describe neural computation at different scales and in different forms. The growing diversity of explanatory objects raises a more basic question: what does a given result establish about a mechanism, and what remains to be resolved?

\section{Principles for Mechanistic Research}

\label{section3}

As shown in Figure~\ref{fig:main}, we organize how the principles below
organize mechanistic research into a recursive logic of inference.
P0 concerns the measurement choices underlying every inference, while
P1 -- P5 address distinct gaps between observations and mechanistic claims.
An individual study may address a subset of these gaps, without following
a fixed sequence.

\textbf{Principle 0: Mechanistic inference is measurement-dependent.} 

Before any mechanistic inference, researchers should always ask how much of the interpretation depends on how it is measured. This point undergirds all five of the principles below.

Empirical studies have shown that measurement matters. Saliency and feature-visualization methods can produce compelling structure that is only weakly tied to the model computation they are intended to explain \citep{adebayo2018sanity,geirhos2024dont}. Activation patching can vary with the corruption procedure and evaluation metric \citep{zhang2024towards}. Different measurement choices can yield different views of the same underlying system. As in other sciences, measurements are not neutral -- what we observe depends partly on the operations and assumptions we use to make them \citep{bechtel1988philosophy,bridgman1927logic} and are therefore part of the evidence for the interpretation.

\textbf{Principle 1: Treat the explanatory object as a hypothesis.} 

In MI studies, researchers typically start with a privileged explanatory object, such as the single neuron, feature, attention head, or circuit. In practice, however, there is a tendency to assume that the phenomenon of interest is \textit{also} organized at the level where the method returned promising results, without considering other levels of analysis.

Evidence from both brains and ANNs shows why this assumption can be misleading. In the prefrontal cortex, task variables remain decodable from population activity even after individually selective neurons are excluded \citep{rigotti2013importance}, indicating that organizational structure can be easily missed at the level of single neurons. A similar pattern appears in ANNs where unit selectivity is sometimes a poor predictor of downstream task performance \citep{morcos2018importance}. Meanwhile, this does not necessarily mean that the single neuron is the wrong object. \citet{bau2020understanding}, for example, found semantically interpretable units whose causal manipulation affected model behavior. Thus, researchers should treat the explanatory object as an empirical hypothesis, not an \emph{a priori} assumption. 

The choice of explanatory object or level of analysis is constrained in two directions. First, it depends on the phenomena being explained. Researchers should specify the explanandum precisely enough to determine what kind of organization an explanation needs to capture. Second, evidence at adjacent levels can test whether the proposed object is sufficient. For example, when a selective unit is found, one can ask whether the relevant computation remains in the population code; conversely, when a population-level structure is identified, one can ask whether localized circuits or neurons are sufficient to explain the behavior. This need not be an exhaustive search over all possible levels.

\textbf{Principle 2: Distinguish represented from computationally implicated information.} 

Once a candidate representation has been identified, a next step is to ask what information it encapsulates. Here, it is important to distinguish between information that is merely represented (total decodable information) versus information that is used by downstream computations (actual decoded information). Decodability is often used as evidence for the former. However, a variable may be decodable from an internal representation without playing a functional role in producing the behavior \citep{diedrichsen2017representational, ritchie2019decoding, kriegeskorte2019interpreting}. 

This distinction has been demonstrated empirically. In human vision, information of visual category can be decoded across broad parts of the visual system, yet only a subset of these representations predicts categorization behavior \citep{grootswagers2018finding}. A similar dissociation appears in ANNs. Models can encode linguistic properties even when those properties are not needed for the task on which the model was trained \citep{ravichander2021probing}, and probing accuracy does not necessarily track the importance of that information for model behavior \citep{elazar2021amnesic}. Thus, finding that a variable is represented identifies information available in the system, but leaves open whether and where that information enters the computation.

The practical upshot is that researchers should explicitly link information identified upstream with downstream states or behavior. Most commonly, one can selectively remove or change the information to see if the downstream computations change as expected.

\textbf{Principle 3: Distinguish causal relevance and causal role.} 

Interventions move mechanistic inference from asking whether information is present to asking how the system depends on it. Ablating, patching, or steering a component can establish that changing it affects the behavior of interest. But the same intervention effect may be consistent with different accounts of what that component does.

Neuroscience provides a clear example. In zebra finches, acute silencing of a song region disrupts singing immediately, whereas permanent lesions cause only transient deficits that are then recovered. This is because acute interventions disrupt downstream circuits rather than isolate local function \citep{otchy2015acute, young2000imputing}. Similarly in ANNs. After attention heads are ablated, downstream components can partially compensate for the perturbation, and the amount of this self-repair varies across inputs \citep{rushing2024explorations}. Thus, an observed change after intervention reflects both the targeted component and how the rest of the system responds to the manipulation. 

Identifying causal role requires designing interventions that can distinguish between competing explanations of what a component does. Researchers should state the operation a component is hypothesized to perform and derive predictions that follow from that account. For example, under perturbations, one can ask where and when the effect should appear, which downstream quantities should change and how, and which should remain relatively unaffected.

\textbf{Principle 4: Distinguish causal roles from mechanistic organization.} 

The next step in elevating components and their causal roles in target computations to the status of `mechanism' is to explore the dependencies among those roles. This step shifts the focus from individual component-role mappings to the relations among them.

Neuroscience provides an example in the case of motor planning in the mouse's anterior lateral motor cortex (ALM). During a delayed-response task, neurons in both ALM hemispheres exhibit preparatory activity that predicts the upcoming movement. There are three possibilities for such preparatory activity: it depends on local recurrent dynamics within one hemisphere, it is maintained redundantly across the two hemispheres, or the two hemispheres depend on each other. \citet{li2016robust} distinguished among them by comparing the results of different perturbations. When one ALM hemisphere was transiently silenced, the population activity was displaced, but after the perturbation ended, the activity pattern rapidly recovered. Meanwhile, bilateral silencing showed that activity did not recover by the end of the delay period. The contrasting outcomes of the two perturbations therefore revealed that the computation was not simply tied to the uninterrupted activity of a particular local population. Rather, these results supported an organization of selectively coupled and partially redundant modules where either hemisphere could maintain the preparatory dynamics, while coupling between them allowed activity in one module to restore the other after a transient perturbation.

This suggests a practical strategy for moving from causal roles to mechanistic organization. Once several components have been assigned candidate roles, researchers can ask what possible dependencies must hold between them. Then, one can perform a series of perturbations on particular inputs, outputs, or paths and measure the intermediate quantities predicted to depend on them.

\textbf{Principle 5: Recurse to search for invariants and revisit the explanatory object.} 

The organization recovered above provides evidence about the question which began the analysis: whether the original explanatory object was the appropriate level. A level that appeared natural when individual components were first identified might turn out to capture implementation-specific details, but the stable organization might lie elsewhere \citep{marr1982vision}. For example, circuit-level algorithms remained relatively consistent across training, task, and model scale even when the specific components implementing them change \citep{tigges2024llm, merullo2024circuit}. Similarly, across independently trained SAEs, individual features may be unstable while the subspaces occupied remain reproducible \citep{gerasimov2026unstable}. Thus, an explanation might remain stable and generalize at a higher level of computation while changing at the lower level of individual components. 

In practice, a recursive approach can potentially reveal which aspects of a mechanism are stable by testing how its validity changes across conditions. The result might support the original level, require a more abstract description, or reveal that several levels capture different parts of the mechanism. Therefore, the principles are better viewed as recursively applied than as a static checklist.

\section{A Historical Warning: When Interpretable Units Become Mechanisms}
\label{section4}

A look to the history of neuroscience shows how the distinctions between the five principles emerged.

Early work asked what a selective neuron (e.g., one responding to an oriented bar or a face) in the brain explained. Gnostic cells and face cells were initially offered as an answer to the question of how objects are represented \citep{barlow1972single, konorski1967integrative, gross1972visual}. The idea of population coding (i.e., distributed representations) shifted the focus from the tuning of individual neurons to the information carried jointly by many cells \citep{georgopoulos1986neuronal, haxby2001distributed, kriegeskorte2008representational}. Dynamical system accounts ask how population activity evolves over time to explain how the structure of the information within the population is made accessible to downstream neural populations that read that information out to support behavior \citep{churchland2012neural,mante2013context,sussillo2013dynamics}
. Progress was made as a broad question such as ``how does the brain recognize objects'' was progressively factored into a set of more precise mechanistic questions: what latent factors are represented, how is information organized, how is it transformed, and how does it ultimately support readout to guide vision-dependent behaviors? Notice that each question admits a different object of description (P1) and its own evidence that the represented information is read out (P2). The notion of `causal' evidence (P3) also went through waves of refinement. Brain lesions, pharmacological inactivations, or microstimulation studies could establish that selective neurons like those for faces were indeed causally involved in face perception behaviors \citep{afraz2006microstimulation, sadagopan2017causal, moeller2017effect, parvizi2012electrical}. However, knowing that a region is causally involved still leaves open the questions of \emph{what} computation it performs and \emph{how} downstream regions interact to support this behavior (P4). Answering these questions requires explicit models and carefully designed experiments \citep{chang2017code}.

Despite these precedents from neuroscience, interpretability work on convolutional neural network (CNN) models of computer vision has followed a remarkably similar historical path. Early visualization studies revealed a progression from lower-level visual features to more complex patterns across CNN layers \citep{zeiler2014visualizing}, and individual units within layers were found to detect objects and human-interpretable concepts \citep{zhou2014object,bau2017network}. But unit interpretability was basis-dependent, and selectivity was later shown to be a poor predictor of task importance \citep{morcos2018importance, dipani2026category}. At the same time, causal interventions demonstrated that some interpretable units do have functional roles \citep{bau2018gan,bau2020understanding}. Again, selectivity (P1) and causal contribution (P2 and 3) answer different questions, and neither alone determines the full organization of the computation (P4). The shared historical pattern across neuroscience and MI points to a general challenge in mechanistic research. An object revealed by an interpretability method should not automatically be treated as a `mechanism' in the scientific or computational sense. The five principles make explicit the required inferential steps, bridging from what an interpretability method reveals to valid mechanistic claims built on that evidence.

\section{What Comes Next: Mechanistic Interpretability as a Science of Discovery}
\label{section5}

MI should be viewed as a science of discovery in two distinct senses. 

First, discovering how a neural network represents information, performs computations, and produces behavior itself constitutes scientific discovery. The historical parallel between neuroscience and MI illustrates that progress requires repeatedly refining the explanatory object and the inferences drawn from each class of evidence. Thus, for MI to become a cumulative science of such mechanisms, empirical findings must be connected to mechanistic claims through a shared logic of inference. Our five principles are intended to provide the structure for making such inference validly. They locate what a result establishes, what remains unresolved, and which mechanistic question follows.

The second sense of scientific discovery is using models as instruments to learn about domains beyond themselves. In this case, models trained on data from a target domain are then analyzed by MI techniques to extract candidate mechanisms about the external domain \citep{simon2025interplm}. Here, recovering the computation or mechanism of a model is an intermediate step because a science of model mechanisms is not automatically a science of models of the external world. A mechanism found in the model is, first, a source of hypotheses about that domain. The gap between an MI mechanism and a scientific hypothesis exists because the model might exploit representations or computations that differ from those that generated its training data, so correspondence cannot be assumed. The five principles of our framework also explain what a model mechanism can contribute to scientific discovery beyond the model by turning interpretability results into a precise candidate hypothesis and set of predictions. These predictions can be tested independently of the interpretability analysis and, where possible, independently of the model itself. Behavioral experiments, new datasets, interventions, or domain-specific measurements can determine whether the identified structure reflects the model alone or captures a general regularity of a scientific domain.

In summary, the five principles of our framework guide the potential contributions of MI to scientific discovery along two linked cycles. First, they describe a recursive and rigorous science of mechanisms within models. Second, they serve within a broader scientific cycle in which these mechanisms generate testable hypotheses about the scientific domains the models capture. Although mechanisms found in models can expose candidate explanations, subsequent independent evidence ultimately determines what they reveal beyond the model and of the external world.

\section{Responsible Use} 
Mechanistic interpretations may be overgeneralized beyond the evidence supporting them, particularly when model mechanisms are taken to reflect mechanisms in an external scientific domain. We emphasize that model-derived mechanisms should generate testable hypotheses rather than substitute for independent, domain-specific empirical validation.

\begin{ack}
We would like to sincerely thank Professor Guillaume Lajoie from Mila and the University of Montreal for his invaluable guidance, thoughtful suggestions, and continuous support. We also express our deep gratitude to Zhiyue Han from the University of Washington-Seattle for insightful discussions and valuable feedback throughout this project.
\end{ack}

\newpage
\bibliographystyle{plainnat}
\bibliography{nips}

%%%%%%%%%%%%%%%%%%%%%%%%%%%%%%%%%%%%%%%%%%%%%%%%%%%%%%%%%%%%

\appendix

%%%%%%%%%%%%%%%%%%%%%%%%%%%%%%%%%%%%%%%%%%%%%%%%%%%%%%%%%%%%

%\newpage
%\input{checklist.tex}

\end{document}